%% file: paper.tex
\documentclass[11pt,letterpaper]{article}
\usepackage[letterpaper]{geometry}
\usepackage{acl2012}
\usepackage{times}
\usepackage{latexsym}
\usepackage{amsmath}
\usepackage{multirow}
\usepackage{url}
\makeatletter
\newcommand{\@BIBLABEL}{\@emptybiblabel}
\newcommand{\@emptybiblabel}[1]{}
\makeatother
\usepackage[hidelinks]{hyperref}
\usepackage{amsfonts}

\usepackage{color}
\usepackage{graphicx}
\usepackage{textcomp}
\usepackage{fancyvrb}
\usepackage{umoline}
\usepackage{footnote}
\usepackage{tabularx, booktabs}
\usepackage{soul}
\usepackage{enumitem}
\usepackage{algpseudocode}
\usepackage{algcompatible}
\usepackage{subcaption}

\usepackage{pifont}
\newcommand{\Yes}{\checkmark}
\newcommand{\No}{$\times$}

\newcolumntype{Y}{>{\centering\arraybackslash}X}

\newcommand{\com}[1]{}
\newcommand{\tb}[1]{\textbf{#1}}
\newcommand{\mb}[1]{\mathbf{#1}}
\newcommand{\be}{\begin{equation}}
\newcommand{\ee}{\end{equation}}

\definecolor{green}{rgb}{0.1,0.5,0.1}
\definecolor{red}{rgb}{1.0,0.2,0.2}
\definecolor{blue}{rgb}{0.0,0.0,1.0}



\makeatletter
\renewcommand{\paragraph}{%
  \@startsection{paragraph}{4}%
  {\z@}{1.0ex \@plus .1ex \@minus .2ex}{-2em}%
  {\normalfont\normalsize\bfseries}%
}
\makeatother

\newcommand{\specialcell}[2]{%
  \begin{tabular}[#1]{@{}#1@{}}#2\end{tabular}}

\title{Aspect-augmented Adversarial Networks for Domain Adaptation}

\author{Yuan Zhang, Regina Barzilay, and Tommi Jaakkola\\
	    Computer Science and Artificial Intelligence Laboratory\\
	    Massachusetts Institute of Technology\\
	    {\tt \{yuanzh, regina, tommi\}@csail.mit.edu}
	  }

\date{}

\begin{document}
\maketitle
\begin{abstract}
\input{abstract}
\end{abstract}

\input{new-intro}
\input{related}
\input{model}
\input{setup}
\input{results}

\input{conclusion}

\section*{Acknowledgments}
The authors acknowledge the support of the U.S. Army Research Office under grant number W911NF-10-1-0533. We thank the MIT NLP group and the TACL reviewers for their comments. Any opinions, findings, conclusions, or recommendations expressed in this paper are those of the authors, and do not necessarily reflect the views of the funding organizations.

\bibliographystyle{acl2012}
\bibliography{paper}

\end{document}

%% file: abstract.tex

We introduce a neural method for transfer learning between two (source and target) classification tasks or aspects over the same domain. Rather than training on target labels, we use a few keywords pertaining to source and target aspects indicating sentence relevance instead of document class labels. Documents are encoded by learning to embed and softly select relevant sentences in an aspect-dependent manner. A shared classifier is trained on the source encoded documents and labels, and applied to target encoded documents. We ensure transfer through aspect-adversarial training so that encoded documents are, as sets, aspect-invariant.
Experimental results demonstrate that our approach outperforms different baselines and model variants on two datasets, yielding an improvement of 27\% on a pathology dataset and 5\% on a review dataset.\footnote{The code is available at \url{https://github.com/yuanzh/aspect_adversarial}.}

%% file: new-intro.tex
\section{Introduction}

\begin{figure}[t]
\centering
 \includegraphics[width=0.48\textwidth]{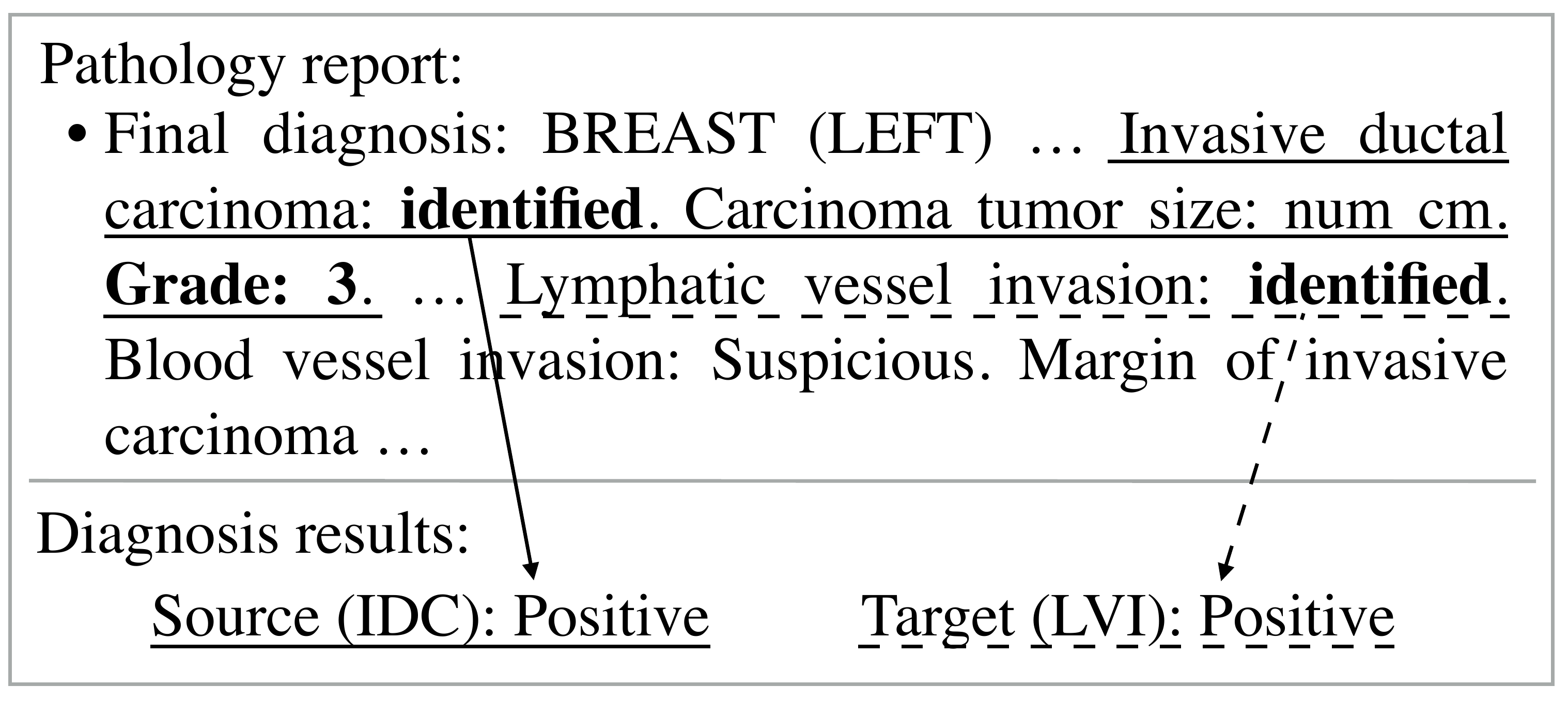}
\caption{A snippet of a breast pathology report with diagnosis results for two types of disease (aspects): carcinoma (IDC) and lymph invasion (LVI). Note how the same phrase indicating positive results (e.g. \emph{identified}) is applicable to both aspects.
A transfer model learns to map other key phrases (e.g. \emph{Grade 3}) to such shared indicators.
}\label{fig:pathology}
\end{figure}

Many NLP problems are naturally multitask classification problems. For instance, values extracted for different fields from the same document are often dependent as they share the same context. Existing systems rely on this dependence (transfer across fields) to improve accuracy. In this paper, we consider a version of this problem where there is a clear dependence between two tasks but annotations are available only for the source task. For example, the target goal may be to classify pathology reports (shown in Figure \ref{fig:pathology}) for the presence of lymph invasion but training data are available only for carcinoma in the same reports. We call this problem \emph{aspect transfer} as the objective is to learn to classify examples differently, focusing on different aspects, without access to target aspect labels. Clearly, such transfer learning is possible only with auxiliary information relating the tasks together. 

The key challenge is to articulate and incorporate commonalities across the tasks. For instance, in classifying reviews of different products, sentiment words (referred to as \emph{pivots}) can be shared across the products. This commonality enables one to align  feature spaces across multiple products, enabling useful transfer~\cite{blitzer2006domain}. Similar properties hold in other contexts and beyond sentiment analysis. Figure~\ref{fig:pathology} shows that certain words and phrases like ``identified'', which indicates the presence of a histological property, are applicable to both carcinoma and lymph invasion. Our method learns and relies on such shared indicators, and utilizes them for effective transfer. 

The unique feature of our transfer problem is that both the source and the target classifiers operate over the same domain, i.e., the same examples. In this setting, traditional transfer methods will always predict the same label for both aspects and thus leading to failure. Instead of supplying the target classifier with direct training labels, our approach builds on a secondary relationship between the tasks using aspect-relevance annotations of sentences. These relevance annotations indicate a possibility that the answer could be found in a sentence, not what the answer is. One can often write simple keyword rules that identify sentence relevance to a particular aspect through representative terms, e.g., specific hormonal markers in the context of pathology reports. Annotations of this kind can be readily provided by domain experts, or extracted from medical literature such as codex rules in pathology~\cite{pantanowitz2008informatics}. We assume a small number of relevance annotations (rules) pertaining to both source and target aspects as a form of weak supervision. We use this sentence-level aspect relevance to learn how to encode the examples (e.g., pathology reports) from the point of view of the desired aspect. In our approach, we construct different aspect-dependent encodings of the same document by softly selecting sentences relevant to the aspect of interest. The key to effective transfer is how these encodings are aligned. 

This encoding mechanism brings the problem closer to the realm of standard domain adaptation, where the derived aspect-specific representations are considered as different domains. Given these representations, our method learns a label classifier shared between the two domains. To ensure that it can be adjusted only based on the source class labels, and that it also reasonably applies to the target encodings, we must align the two sets of encoded examples.\footnote{This alignment or invariance is enforced on the level of sets, not individual reports; aspect-driven encoding of any specific report should remain substantially different for the two tasks since the encoded examples are passed on to the same classifier.}  Learning this alignment is possible because, as discussed above, some keywords are directly transferable and can serve as anchors for constructing this invariant space. To learn this invariant representation, we introduce an adversarial domain classifier analogous to the recent successful use of adversarial training in computer vision~\cite{ganin2014unsupervised}. The role of the domain classifier (adversary) is to learn to distinguish between the two types of encodings. During training we update the encoder with an adversarial objective to cause the classifier to fail.  
The encoder therefore learns to eliminate aspect-specific information so that encodings look invariant (as sets) to the classifier, thus establishing aspect-invariance encodings and enabling transfer.
All three components in our approach, 1) aspect-driven encoding, 2) classification of source labels, and 3) domain adversary, are trained jointly (concurrently) to complement and balance each other.

Adversarial training of domain and label classifiers can be challenging to stabilize. In our setting, sentences are encoded with a convolutional model.
Feedback from adversarial training can be an unstable guide for how the sentences should be encoded.
To address this issue, we incorporate an additional word-level auto-encoder reconstruction loss to ground the convolutional processing of sentences. We empirically demonstrate that this additional objective yields richer and more diversified feature representations, improving transfer. 

We evaluate our approach on pathology reports (aspect transfer) as well as on a more standard review dataset (domain adaptation). On the pathology dataset, we explore cross-aspect transfer across different types of breast disease. Specifically, we test on six adaptation tasks, consistently outperforming all other baselines. Overall, our full model achieves 27\% and 20.2\% absolute improvement arising from aspect-driven encoding and adversarial training respectively. Moreover, our unsupervised adaptation method is only 5.7\% behind the accuracy of a supervised target model. On the review dataset, we test adaptations from hotel to restaurant reviews.
Our model outperforms the marginalized denoising autoencoder~\cite{chen2012marginalized} by 5\%.
Finally, we examine and illustrate the impact of individual components on the resulting performance. 

%% file: related.tex

\section{Related Work}

\paragraph{Domain Adaptation for Deep Learning}


Existing approaches commonly induce abstract representations without pulling apart different aspects in the same example, and therefore are likely to fail on the aspect transfer problem.
The majority of these prior methods first learn a task-independent representation, and then train a label predictor (e.g. SVM) on this representation in a separate step. 
For example, earlier researches employ a shared autoencoder~\cite{glorot2011domain,chopra2013dlid} to learn a cross-domain representation.  
\newcite{chen2012marginalized} further improve and stabilize the representation learning by utilizing marginalized denoising autoencoders.
Later, \newcite{zhou2016bi} propose to minimize domain-shift of the autoencoder in a linear data combination manner.
Other researches have focused on learning transferable representations in an end-to-end fashion. Examples include using transduction learning for object recognition~\cite{sener2016learning} and using residual transfer networks for image classification~\cite{long2016unsupervised}. In contrast, we use adversarial training to encourage learning domain-invariant features in a more explicit way. Our approach offers another two advantages over prior work. First, we jointly optimize features with the final classification task while many previous works only learn task-independent features using autoencoders. Second, our model can handle traditional domain transfer as well as aspect transfer, while previous methods can only handle the former.

\paragraph{Adversarial Learning in Vision and NLP}
Our approach closely relates to the idea of domain-adversarial training. Adversarial networks were originally developed for image generation \cite{goodfellow2014generative,makhzani2015adversarial,springenberg2015unsupervised,radford2015unsupervised,taigman2016unsupervised}, and were later applied to domain adaptation in computer vision~\cite{ganin2014unsupervised,ganin2015domain,domain_separation_nets16,tzeng2014deep} and speech recognition~\cite{shinohara2016adversarial}. 
The core idea of these approaches is to promote the emergence of invariant image features by optimizing the feature extractor as an adversary against the domain classifier.
While \newcite{ganin2015domain} also apply this idea to sentiment analysis, their practical gains have remained limited. 

Our approach presents two main departures. In computer vision, adversarial learning has been used for transferring across domains, while our method can also handle aspect transfer.  In addition, we introduce a
reconstruction loss which results in more robust adversarial training. We believe that this formulation will benefit other applications of adversarial training, beyond the ones described in this paper.

\paragraph{Semi-supervised Learning with Keywords} In our work, we use a small set of keywords as a source of weak supervision for aspect-relevance scoring. This relates to prior work on utilizing prototypes and seed words in semi-supervised learning~\cite{haghighi2006prototype,grenager2005unsupervised,chang2007guiding,mann2008generalized,jagarlamudi2012incorporating,li2012wiki,eisenstein2017unsupervised}. 
All these prior approaches utilize prototype annotations primarily targeting model bootstrapping but not for learning representations. In contrast, our model uses provided keywords to learn aspect-driven encoding of input examples. 



\paragraph{Attention Mechanism in NLP} One may view our aspect-relevance scorer as a sentence-level ``semi-supervised attention'', in which relevant sentences receive more attention during feature extraction. While traditional attention-based models typically induce attention in an unsupervised manner, they have to rely on a large amount of labeled data for the target task~\cite{bahdanau2014neural,rush2015neural,chen2015abc,cheng2016long,xu2015show,xu2015ask,yang2015stacked,martins2016softmax,lei2016rationalizing}. Unlike these methods, our approach assumes no label annotations in the target domain. Other researches have focused on utilizing human-provided rationales as ``supervised attention'' to improve prediction~\cite{zaidan2007using,marshall2015robotreviewer,zhang2016rationale,brun-perez-roux:2016:SemEval}. In contrast, our model only assumes access to a small set of keywords as a source of weak supervision. Moreover, all these prior approaches focus on in-domain classification. In this paper, however, we study the task in the context of domain adaptation. 

\paragraph{Multitask Learning}  Existing multitask learning methods
focus on the case where supervision is available for all tasks. A typical architecture involves using a shared encoder with a separate classifier for each task. \cite{caruana1998multitask,pan2010survey,collobert2008unified,liu2015representation,bordes2012joint}. In contrast, our work assumes labeled data only for the source aspect. We train a single classifier for both aspects by learning aspect-invariant representation that enables the transfer. 

%% file: model.tex
\section{Problem Formulation}

We begin by formalizing \emph{aspect transfer} with the idea of differentiating it from standard domain adaptation. In our setup, we have two classification tasks called the source and the target tasks. In contrast to source and target tasks in domain adaptation, both of these tasks are defined over the same set of examples (here documents, e.g., pathology reports). What differentiates the two classification tasks is that they pertain to different aspects in the examples. If each training document were annotated with both the source and the target aspect labels, the problem would reduce to multi-label classification. However, in our setting training labels are available only for the source aspect so the goal is to solve the target task without any associated training label. 

To fix notation, let $\mb{d}=\{\mb{s}_i\}_{i=1}^{|\mb{d}|}$ be a document that consists of a sequence of $|\mb{d}|$ sentences $\mb{s}_i$. Given a document $\mb{d}$, and the aspect of interest, we wish to predict the corresponding aspect-dependent class label $y$ (e.g., $y\in\{-1,1\}$). We assume that the set of possible labels are the same across aspects. We use $y^s_{l;k}$ to denote the k-th coordinate of a one-hot vector indicating the correct training source aspect label for document $\mb{d}_l$. Target aspect labels are not available during training. 


Beyond labeled documents for the source aspect $\{\mb{d}_l,y^s_l\}_{l\in L}$, and shared unlabeled documents for source and target aspects $\{\mb{d}_l\}_{l\in U}$, we assume further that we have relevance scores pertaining to each aspect. The relevance is given per sentence, for some subset of sentences across the documents, and indicates the possibility that the answer for that document would be found in the sentence but without indicating which way the answer goes. Relevance is always aspect dependent yet often easy to provide in the form of simple keyword rules. 

We use $r^a_{i} \in \{0,1\}$ to denote the given relevance label pertaining to aspect $a$ for sentence $\mb{s}_i$. Only a small subset of sentences in the training set have associated relevance labels. Let $R = \{(a,l,i)\}$ denote the index set of relevance labels such that if $(a,l,i)\in R$ then aspect $a$'s relevance label $r^a_{l,i}$ is available for the $i^{th}$ sentence in document $\mb{d}_l$. In our case relevance labels arise from aspect-dependent keyword matches. $r^a_{i}=1$ when the sentence contains any keywords pertaining to aspect $a$ and $r^a_i=0$ if it has any keywords of other aspects.\footnote{$r^a_i=1$ if the sentence contains keywords pertaining to both aspect $a$ and other aspects.} Separate subsets of relevance labels are available for each aspect as the keywords differ.

The transfer that is sought here is between two tasks over the same set of examples rather than between two different types of examples for the same task as in standard domain adaptation. However, the two formulations can be reconciled if full relevance annotations are assumed to be available during training and testing. In this scenario, we could simply lift the sets of relevant sentences from each document as new types of documents. The goal would be then to learn to classify documents of type $\mathcal{T}$ (consisting of sentences relevant to the target aspect) based on having labels only for type $\mathcal{S}$ (source) documents, a standard domain adaptation task. Our problem is more challenging as the aspect-relevance of sentences must be learned from limited annotations.

Finally, we note that the aspect transfer problem and the method we develop to solve it work the same even when source and target documents are a priori different, something we will demonstrate later.

\begin{figure*}[t]
\centering
\includegraphics[width=0.95\textwidth]{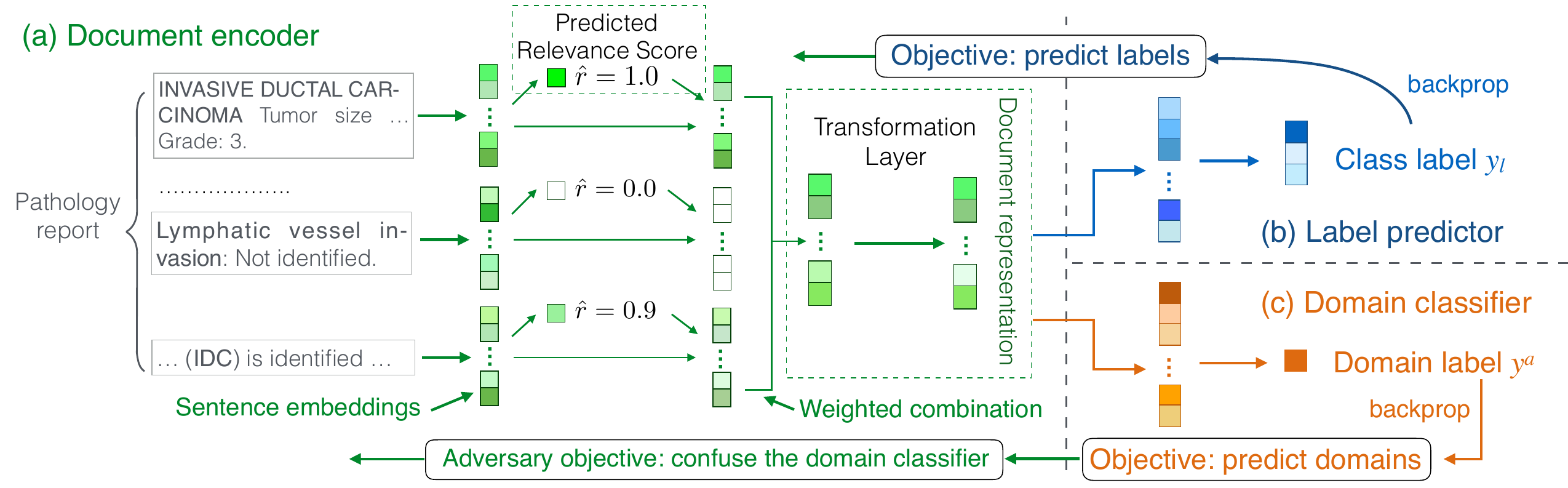}
\caption{Aspect-augmented adversarial network for transfer learning.
The model is composed of (a) an aspect-driven document encoder, (b) a label predictor and (c) a domain classifier. 
}
\label{fig:model}
\end{figure*}

\section{Methods}

\subsection{Overview of our approach}

Our model consists of three key components as shown in Figure \ref{fig:model}. Each document is encoded in a relevance weighted, aspect-dependent manner (green, left part of Figure \ref{fig:model}) and classified using the label predictor (blue, top-right). During training, the encoded documents are also passed on to the domain classifier (orange, bottom-right). The role of the domain classifier, as the adversary, is to ensure that the aspect-dependent encodings of documents are distributionally matched. This matching justifies the use of the same end-classifier to provide the predicted label regardless of the task (aspect).  

To encode a document, the model first maps each sentence into a vector and then passes the vector to a scoring network to determine whether the sentence is relevant for the chosen aspect. These predicted relevance scores are used to obtain document vectors by taking relevance-weighted sum of the associated sentence vectors. Thus, the manner in which the document vector is constructed is always \emph{aspect-dependent} due to the chosen relevance weights.

During training, the resulting adjusted document vectors are consumed by the two classifiers. The primary label classifier aims to predict the source labels (when available), while the domain classifier determines whether the document vector pertains to the source or target aspect, which is the label that we know by construction. Furthermore, we jointly update the document encoder with a reverse of the gradient from the domain classifier, so that the encoder learns to induce document representations that fool the domain classifier. The resulting encoded representations will be aspect-invariant, facilitating transfer. 

Our adversarial training scheme uses all the training losses concurrently to adjust the model parameters. During testing, we simply encode each test document in a target-aspect dependent manner, and apply the same label predictor. We expect that the same label classifier does well on the target task since it solves the source task, and operates on relevance-weighted representations that are matched across the tasks. 
While our method is designed to work in the extreme setting that the examples for the two aspects are the same, this is by no means a requirement. Our method will also work fine in the more traditional domain adaptation setting, which we will demonstrate later.

\begin{figure}[t]
\centering
\includegraphics[width=0.44\textwidth]{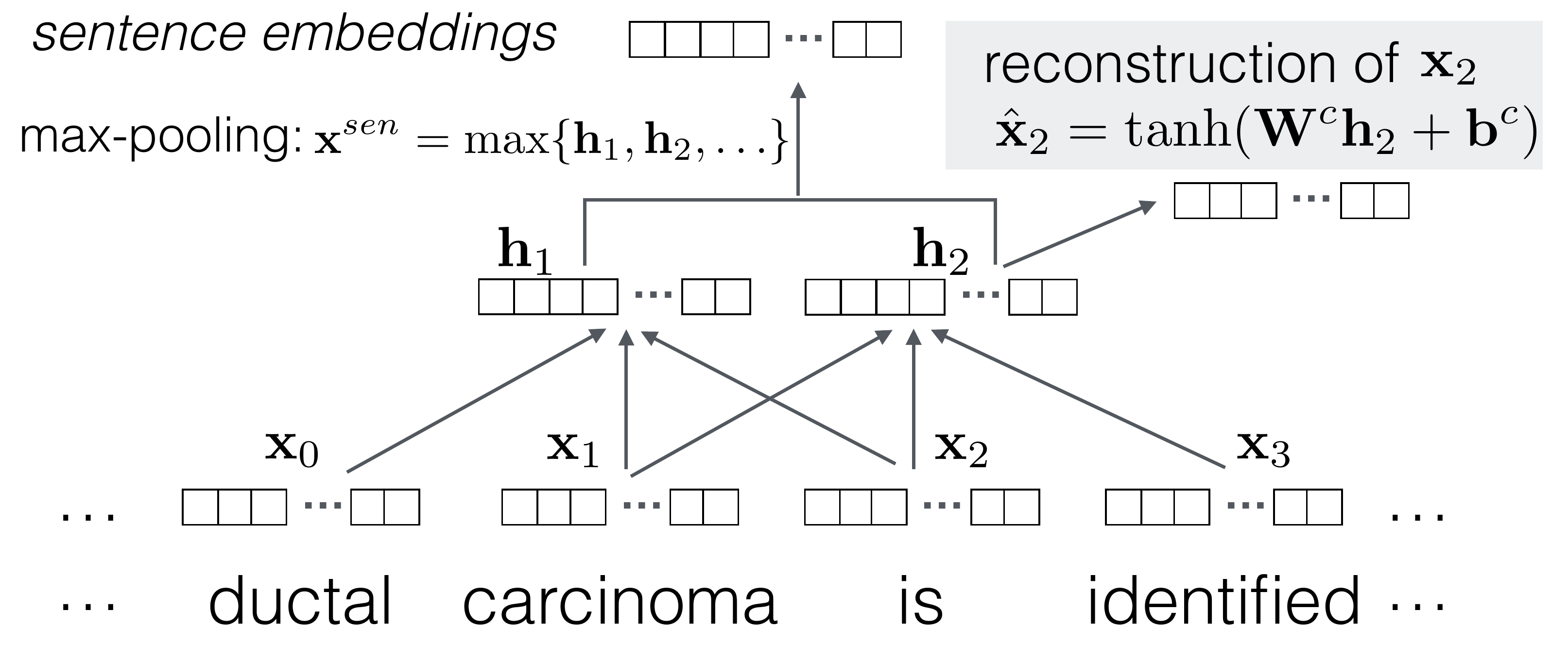}
\caption{Illustration of the convolutional model and the reconstruction of word embeddings from the associated convolutional layer.}
\label{fig:recon}
\end{figure}

\subsection{Components in detail}


\paragraph{Sentence embedding} 
We apply a convolutional model illustrated in Figure \ref{fig:recon} to each sentence $\mb{s}_i$ to obtain sentence-level vector embeddings $\mb{x}^{sen}_i$. The use of RNNs or bi-LSTMs would result in more flexible sentence embeddings but based on our initial experiments, we did not observe any significant gains over the simpler CNNs. 


We further ground the resulting sentence embeddings by including an additional word-level reconstruction step in the convolutional model. The purpose of this reconstruction step is to balance adversarial training signals propagating back from the domain classifier. Specifically, it forces the sentence encoder to keep rich word-level information in contrast to adversarial training that seeks to eliminate aspect specific features. We provide an empirical analysis of the impact of this reconstruction in the experiment section (Section \ref{sec:analysis}).

More concretely, we reconstruct word embedding from the corresponding convolutional layer, as shown in Figure \ref{fig:recon}.\footnote{
This process is omitted in Figure \ref{fig:model} for brevity.} 
We use $\mb{x}_{i,j}$ to denote the embedding of the $j$-th word in sentence $\mb{s}_i$.
Let $\mb{h}_{i,j}$ be the convolutional output when $\mb{x}_{i,j}$ is at the center of the window. We reconstruct $\mb{x}_{i,j}$ by
\be
\hat{\mb{x}}_{i,j} = \tanh(\mb{W}^c\mb{h}_{i,j}+\mb{b}^c)
\ee
where $\mb{W}^c$ and $\mb{b}^c$ are parameters of the reconstruction layer. The loss associated with the reconstruction for document $\mb{d}$ is 
\be
\mathcal{L}^{rec}(\mb{d}) = \frac{1}{n}\sum_{i,j} ||\hat{\mb{x}}_{i,j}-\tanh(\mb{x}_{i,j})||_2^2
\ee
where $n$ is the number of tokens in the document and indexes $i$, $j$ identify the sentence and word, respectively. The overall reconstruction loss $\mathcal{L}^{rec}$ is obtained by summing over all labeled/unlabeled documents.

\paragraph{Relevance prediction} 
We use a small set of keyword rules to generate binary relevance labels, both positive ($r=1$) and negative ($r=0$). These labels represent the only supervision available to predict relevance. The prediction is made on the basis of the sentence vector $\mb{x}^{sen}_i$ passed through a feed-forward network with a ReLU output unit. The network has a single shared hidden layer and a separate output layer for each aspect. Note that our relevance prediction network is trained as a non-negative regression model even though the available labels are binary, as relevance varies more on a linear rather than binary scale.

Given relevance labels indexed by $R=\{(a,l,i)\}$, we minimize 
\be
\mathcal{L}^{rel} = \sum_{(a,l,i)\in R} \big(r^a_{l,i} - \hat r^a_{l,i}\big)^2
\ee
where $\hat r^a_{l,i}$ is the predicted (non-negative) relevance score pertaining to aspect $a$ for the $i^{th}$ sentence in document $\mb{d}_l$, as shown in the left part of Figure~\ref{fig:model}. $r^a_{l,i}$, defined earlier, is the given binary (0/1) relevance label. 
We use a score in $[0, 1]$ scale because it can be naturally used as a weight for vector combinations, as shown next.

\paragraph{Document encoding}

The initial vector representation for each document such as $\mb{d}_l$ is obtained as a relevance weighted combination of the associated sentence vectors, i.e.,
\be
\mb{x}^{doc,a}_l = 
\frac{\sum_{i}\hat{r}^a_{l,i}\cdot \mb{x}^{sen}_{l,i}}{\sum_{i}\hat{r}^a_{l,i}}
\ee
The resulting vector selectively encodes information from the sentences based on relevance to the focal aspect.

\paragraph{Transformation layer} The manner in which document vectors arise from sentence vectors means that they will retain aspect-specific information that will hinder transfer across aspects. To help remove non-transferable information, we add a transformation layer to map the initial document vectors $\mb{x}^{doc,a}_l$ to their domain invariant (as a set) versions, as shown in Figure \ref{fig:model}. Specifically, the transformed representation is given by $\mb{x}^{tr,a}_l=\mb{W}^{tr}\mb{x}^{doc,a}_l$. Meanwhile, the transformation has to be strongly regularized lest the gradient from the adversary would wipe out all the document signal.  
We add the following regularization term 
\be\label{eq:transformation}
\Omega^{tr} = \lambda^{tr}||\mb{W}^{tr}-\mb{I}||_F^2
\ee
to discourage significant deviation away from identity $\mb{I}$. $\lambda^{tr}$ is a regularization parameter that has to be set separately based on validation performance. We show an empirical analysis of the impact of this transformation layer in Section~\ref{sec:analysis}.

\paragraph{Primary label classifier} As shown in the top-right part of Figure~\ref{fig:model}, the classifier takes in the adjusted document representation as an input and predicts a probability distribution over the possible class labels. The classifier is a feed-forward network with a single hidden layer using ReLU activations and a softmax output layer over the possible class labels. 
Note that we train only one label classifier that is shared by both aspects.
The classifier operates the same regardless of the aspect to which the document was encoded. It must therefore be co-operatively learned together with the encodings. 

Let $\hat p_{l;k}$ denote the predicted probability of class $k$ for document $\mb{d}_l$ when the document is encoded from the point of view of the source aspect. Recall that $[y^s_{l;1},\ldots,y^s_{l;m}]$ is a one-hot vector for the correct (given) source class label for document $\mb{d}_l$, hence also a distribution. We use the cross-entropy loss for the label classifier
\be
\mathcal{L}^{lab} = \sum_{l\in L}\left[
-\sum_{k=1}^m y^s_{l;k} \log\hat{p}_{l;k}
\right] 
\ee

\paragraph{Domain classifier} As shown in the bottom-right part of Figure~\ref{fig:model}, the domain classifier functions as an adversary to ensure that the documents encoded with respect to the source and target aspects look the same as sets of examples. The invariance is achieved when the domain classifier (as the adversary) fails to distinguish between the two. Structurally, the domain classifier is a feed-forward network with a single ReLU hidden layer and a softmax output layer over the two aspect labels.

Let $y^a = [y^a_{1},y^a_{2}]$ denote the one-hot domain label vector for aspect $a\in \{s,t\}$. In other words, $y^s = [1,0]$ and $y^t = [0,1]$. We use $\hat q_{k}(\mb{x}^{tr,a}_l)$ as the predicted probability that the domain label is $k$ when the domain classifier receives $\mb{x}^{tr,a}_l$ as the input. The domain classifier is trained to minimize 
\be
\mathcal{L}^{dom} = \sum_{l\in L\cup U}\sum_{a\in \{s,t\}}\left[
-\sum_{k=1}^2 y^a_k \log \hat q_k(\mb{x}^{tr,a}_l)
\right]
\ee

\subsection{Joint learning}

We combine the individual component losses pertaining to word reconstruction, relevance labels, transformation layer regularization, source class labels, and domain adversary into an overall objective function
\begin{align}
&\mathcal{L}^{all} = 
\mathcal{L}^{rec}+\mathcal{L}^{rel}+\Omega^{tr}+\mathcal{L}^{lab}
-\rho\mathcal{L}^{dom}
\end{align}
which is minimized with respect to the model parameters except for the adversary (domain classifier). The adversary is maximizing the same objective with respect to its own parameters. The last term $-\rho\mathcal{L}^{dom}$ corresponds to the objective of causing the domain classifier to fail. The proportionality constant $\rho$ controls the impact of gradients from the adversary on the document representation; the adversary itself is always directly minimizing $\mathcal{L}^{dom}$.

All the parameters are optimized jointly using standard backpropagation (concurrent for the adversary). Each mini-batch is balanced by aspect, half coming from the source, the other half from the target. All the loss functions except $\mathcal{L}^{lab}$ make use of both labeled and unlabeled documents. Additionally, it would be straightforward to add a loss term for target labels if they are available. 


%% file: setup.tex

\section{Experimental Setup}\label{sec:setup}

\begin{table}[t]
\begin{small}
    \centering
    \begin{tabular}{llcc}
	\toprule
    \multicolumn{2}{l}{\textsc{Dataset}} & \#Labeled & \#Unlabeled \\
    \midrule
    \multirow{4}{*}{\textsc{Pathology}} & DCIS & 23.8k & \multirow{4}{*}{96.6k} \\
     & LCIS & 10.7k &  \\
     & IDC & 22.9k &  \\
     & ALH & 9.2k &  \\
     \midrule
    \multirow{2}{*}{\textsc{Review}} & Hotel & 100k & 100k \\
     & Restaurant & - & 200k \\
    \bottomrule
    \end{tabular}
\end{small}
    \caption{Statistics of the pathology reports dataset and the reviews dataset that we use for training. Our model utilizes both labeled and unlabeled data.
    }\label{tb:data}
\end{table}

\paragraph{Pathology dataset}
This dataset contains 96.6k breast pathology reports collected from three hospitals~\cite{Yala:2016a}. A portion of this dataset is manually annotated with 20 categorical values, representing various aspects of breast disease. In our experiments, we focus on four aspects related to carcinomas and atypias:  Ductal Carcinoma In-Situ (DCIS), Lobular Carcinoma In-Situ (LCIS), Invasive Ductal Carcinoma (IDC) and Atypical Lobular Hyperplasia (ALH). Each aspect is annotated using binary labels. We use 500 held out reports as our test set and use the rest of the labeled data as our training set: 23.8k reports for DCIS, 10.7k for LCIS, 22.9k for IDC, and 9.2k for ALH. Table \ref{tb:data} summarizes statistics of the dataset.

\begin{table}[t]
    \centering
    \begin{tabular}{ll}
	\toprule
    \textsc{Aspect} & \textsc{Keywords} \\
    \midrule
    IDC & IDC, Invasive Ductal Carcinoma \\
    ALH & ALH, Atypical Lobular Hyperplasia \\
    \bottomrule
    \end{tabular}
    \caption{Examples of aspects and their corresponding keywords (case insensitive) in the pathology dataset.}\label{tb:keywords}
\end{table}

We explore the adaptation problem from one aspect to another. For example, we want to train a model on annotations of DCIS and apply it on LCIS. 
For each aspect, we use up to three 
common names as a source of supervision for learning the relevance scorer, as illustrated in Table \ref{tb:keywords}. Note that the provided list is by no means  exhaustive. In fact~\newcite{buckley2012feasibility} provide example of 60 different verbalizations of LCIS, not counting negations.

\paragraph{Review dataset} Our second experiment is based on a domain transfer of sentiment classification. For the source domain, we use the hotel review dataset introduced in previous work \cite{wang2010latent,wang2011latent}, and for the target domain, we use the restaurant review dataset from Yelp.\footnote{The restaurant portion of \url{https://www.yelp.com/dataset_challenge}.} Both datasets have ratings on a scale of 1 to 5 stars. 
Following previous work~\cite{blitzer2007biographies}, we label reviews with ratings $>3$ as positive and those with ratings $<3$ as negative, discarding the rest. The hotel dataset includes a total of around 200k reviews collected from TripAdvisor,\footnote{\url{https://www.tripadvisor.com/}} so we split 100k as labeled and the other 100k as unlabeled data. We randomly select 200k restaurant reviews as the unlabeled data in the target domain. Our test set consists of 2k reviews. Table \ref{tb:data} summarizes the statistics of the review dataset.

The hotel reviews naturally have ratings for six aspects, including \emph{value}, \emph{room} quality, \emph{checkin} service, room \emph{service}, \emph{cleanliness} and \emph{location}. We use the first five aspects because the sixth aspect \emph{location} has positive labels for over 95\% of the reviews and thus the trained model will suffer from the lack of negative examples. The restaurant reviews, however, only have single ratings for an \emph{overall} impression. Therefore, we explore the task of adaptation  from each of the five hotel aspects to the restaurant domain. The hotel reviews dataset also provides a total of 280 keywords for different aspects that are generated by the bootstrapping method used in \newcite{wang2010latent}. We use those keywords as supervision for learning the relevance scorer.

\begin{table}[t]
    \centering
    \begin{tabular}{@{~~}lc@{~~}cc@{~~}cc@{~~}}
	\toprule
    \multirow{2}{*}{\textsc{Method}} & \multicolumn{2}{c}{\textsc{Source}} & \multicolumn{2}{c}{\textsc{Target}} & \multirow{2}{*}{\specialcell{c}{Key-\\word}}\\
    \cmidrule(lr){2-3} \cmidrule(l){4-5}
     & Lab. & Unlab. & Lab. & Unlab. \\
    \midrule
    SVM & \Yes & \No & \No & \No & \No\\
    SourceOnly & \Yes & \Yes & \No & \No & \Yes\\
    mSDA & \Yes & \Yes & \No & \Yes & \No\\
    AAN-NA & \Yes & \Yes & \No & \Yes & \Yes\\
    AAN-NR & \Yes & \Yes & \No & \Yes & \No\\
    In-Domain & \No & \No & \Yes & \No & \Yes\\
    AAN-Full & \Yes & \Yes & \No & \Yes & \Yes\\
    \bottomrule
    \end{tabular}
    \caption{Usage of labeled (Lab.), unlabeled (Unlab.) data and keyword rules in each domain by our model and other baseline methods. 
	AAN-* denote our model and its variants.}
	\label{tb:usage}
\end{table}

\begin{table*}[t]
    \centering
    \begin{tabular}{llcccc@{~~}@{~~}c@{~~}@{~~}c@{~~}@{~~}c}
	\toprule
	\multicolumn{2}{c}{\textsc{Domain}} & \multirow{2}{*}{~SVM~} & ~Source & \multirow{2}{*}{~mSDA~} & \multirow{2}{*}{AAN-NA} & \multirow{2}{*}{~AAN-NR} & \multirow{2}{*}{AAN-Full} & \multirow{2}{*}{In-Domain}\\
    \textsc{~Source} & \textsc{Target~} & & Only & & & & \\
    \midrule
    ~LCIS & \multirow{3}{*}{DCIS} & 45.8 & 25.2 & 45.0 & 81.2 & 50.0 & \tb{93.0} & \multirow{3}{*}{96.2} \\
    ~IDC &  & 71.8 & 62.4 & 73.0 & 87.6 & 81.4 & \tb{94.8} & \\
    ~ALH &  & 37.2 & 20.6 & 39.0 & 49.2 & 48.0 & \tb{84.6} & \\
    \midrule
    ~DCIS & \multirow{3}{*}{LCIS} & 73.8 & 75.4 & 76.2 & 89.0 & 81.2 & \tb{95.2} & \multirow{3}{*}{97.8} \\
    ~IDC &  & 71.4 & 66.4 & 71.6 & 84.8 & 52.0 & \tb{85.0} & \\
    ~ALH &  & 54.4 & 46.4 & 54.2 & 84.8 & 52.4 & \tb{93.2} & \\
    \midrule
    ~DCIS & \multirow{3}{*}{IDC} & 94.0 & 77.4 & 94.0 & 92.4 & 93.8 & \tb{95.4} & \multirow{3}{*}{96.8} \\
    ~LCIS &  & 51.6 & 29.5 & 53.2 & 89.6 & 51.2 & \tb{93.8} & \\
    ~ALH &  & 41.0 & 26.8 & 39.2 & 68.0 & 31.6 & \tb{89.6} & \\
	\midrule
    ~DCIS & \multirow{3}{*}{ALH} & 74.6 & 75.0 & 75.0 & 52.6 & 74.2 & \tb{90.4} & \multirow{3}{*}{96.8} \\
    ~LCIS &  & 59.0 & 51.6 & 60.4 & 52.6 & 60.0 & \tb{92.8} &  \\
    ~IDC &  & 67.6 & 66.4 & 68.8 & 52.6 & 69.2 & \tb{87.0} &  \\
    \midrule
    \multicolumn{2}{c}{\textsc{Average}}  & 61.9 & 51.9 & 62.5 & 71.0 & 64.2 & \tb{91.2} & 96.9 \\
	\bottomrule
    \end{tabular}
    \caption{\tb{Pathology:} Classification accuracy (\%) of different approaches on the pathology reports dataset, including the results of twelve adaptation scenarios from four different aspects (IDC, ALH, DCIS and LCIS) in breast cancer pathology reports. ``mSDA'' indicates the marginalized denoising autoencoder in \protect\cite{chen2012marginalized}. ``AAN-NA'' and ``AAN-NR'' corresponds to our model without the adversarial training and the aspect-relevance scoring component, respectively. We also include in the last column the in-domain supervised training results of our model as the performance upper bound. Boldface numbers indicate the best accuracy for each testing scenario.}\label{tb:pathology}
\end{table*}

\paragraph{Baselines and our method} We first compare against a linear \textbf{SVM} trained on the raw bag-of-words representation of labeled data in source. Second, we compare against our \textbf{SourceOnly} model that assumes no target domain data or keywords. It thus has no adversarial training or target aspect-relevance scoring. Next we compare with marginalized Stacked Denoising Autoencoders (\textbf{mSDA})~\cite{chen2012marginalized}, a domain adaptation algorithm that outperforms both prior deep learning and shallow learning approaches.\footnote{We use the publicly available implementation provided by the authors at \url{http://www.cse.wustl.edu/~mchen/code/mSDA.tar}. We use the hyper-parameters from the authors and their models have more parameters than ours.} 

In the rest part of the paper, we name our method and its variants as \textbf{AAN} (\textbf{A}spect-augmented \textbf{A}dversarial \textbf{N}etworks).
We compare against \textbf{AAN-NA} and \textbf{AAN-NR} that are our model variants without adversarial training and without aspect-relevance scoring respectively. Finally we include supervised models trained on the full set of \textbf{In-Domain} annotations as the performance upper bound.
Table \ref{tb:usage} summarizes the usage of labeled and unlabeled data in each domain as well as keyword rules by our model (\textbf{AAN-Full}) and different baselines. Note that our model assumes the same set of data as the AAN-NA, AAN-NR and mSDA methods.









\paragraph{Implementation details}
Following prior work \cite{ganin2014unsupervised}, we gradually increase the adversarial strength $\rho$ and decay the learning rate during training. We also apply batch normalization \cite{ioffe2015batch} on the sentence encoder and apply dropout with ratio 0.2 on word embeddings and each hidden layer activation. We set the hidden layer size to 150 and pick the transformation regularization weight $\lambda^{t}=0.1$ for the pathology dataset and $\lambda^{t}=10.0$ for the review dataset. 

%% file: results.tex
\begin{table*}[t]
    \centering
    \begin{tabular}{l@{~}@{~}lcccc@{~~}@{~~}c@{~~}@{~~}c@{~~}@{~~}c}
	\toprule
	\multicolumn{2}{c}{\textsc{Domain}} & \multirow{2}{*}{~SVM~} & Source & \multirow{2}{*}{~mSDA~} & \multirow{2}{*}{AAN-NA} & \multirow{2}{*}{AAN-NR} & \multirow{2}{*}{AAN-Full} & \multirow{2}{*}{In-Domain}\\
    \textsc{Source} & \textsc{Target} & & Only & & & & \\
    \midrule
    Value & \multirow{5}{*}{\specialcell{c}{Restaurant\\Overall}} & 82.2 & 87.4 & 84.7 & 87.1 & \tb{91.1} & 89.6 & \multirow{5}{*}{93.4} \\
    Room &  & 75.6 & 79.3 & 80.3 & 79.7 & 86.1 & \tb{86.6} &  \\
    Checkin &  & 77.8 & 83.0 & 81.0 & 80.9 & \tb{87.2} & 85.4 & \\
    Service &  & 82.2 & 88.0 & 83.8 & 88.8 & 87.9 & \tb{89.1} &  \\
    Cleanliness &  & 77.9 & 83.2 & 78.4 & 83.1 & \tb{84.5} & 81.4 &  \\
    \midrule
    \multicolumn{2}{c}{\textsc{Average}}  & 79.1 & 84.2 & 81.6 & 83.9 & \tb{87.3} & 86.4 & 93.4 \\
	\bottomrule
    \end{tabular}
    \caption{\tb{Review:} Classification accuracy (\%) of different approaches on the reviews dataset. 
Columns have the same meaning as in Table \ref{tb:pathology}. Boldface numbers indicate the best accuracy for each testing scenario.}\label{tb:review}
\end{table*}



\begin{figure*}[t]
\centering
\includegraphics[width=0.9\textwidth]{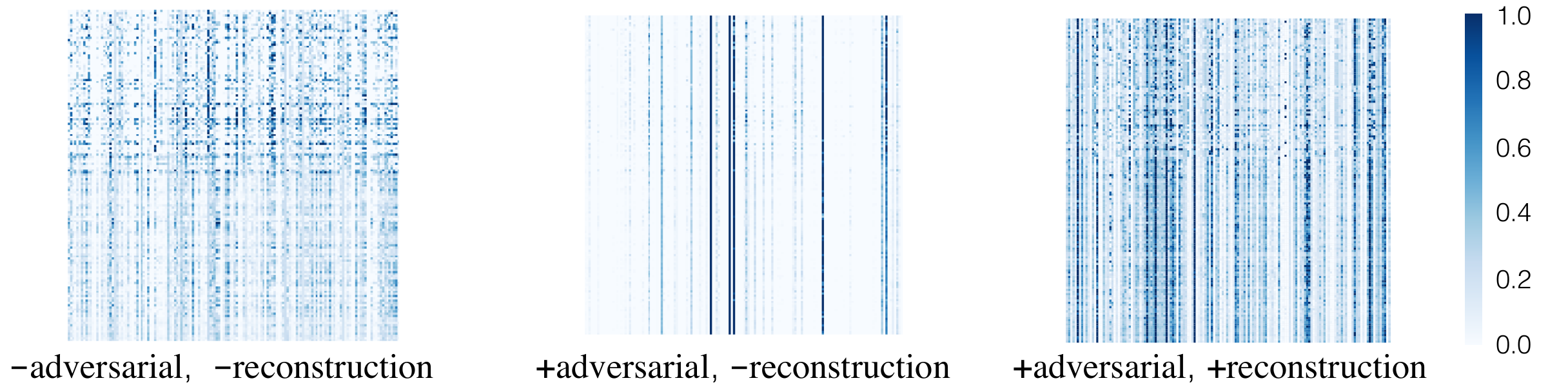}
\caption{Heat map of $150\times 150$ matrices. Each row corresponds to the vector representation of a document that comes from either the source domain (top half) or the target domain (bottom half). Models are trained on the review dataset when room quality is the source aspect. 
}\label{fig:heat}
\end{figure*}

\section{Main Results}\label{sec:main}

Table \ref{tb:pathology} summarizes the classification accuracy of different methods on the pathology dataset, including the results of twelve adaptation tasks. Our full model (AAN-Full) consistently achieves the best performance on each task compared with other baselines and model variants. It is not surprising that SVM and mSDA perform poorly on this dataset because they only predict labels based on an overall feature representation of the input, and do not utilize weak supervision provided by aspect-specific keywords.  As a reference, we also provide a performance upper bound by training our model on the full labeled set in the target domain, denoted as In-Domain in the last column of Table \ref{tb:pathology}. On average, the accuracy of our model (AAN-Full) is only 5.7\% behind this upper bound.

Table \ref{tb:review} shows the adaptation results from each aspect in the hotel reviews to the overall ratings of restaurant reviews. AAN-Full and AAN-NR are the two best performing systems on this review dataset, attaining around 5\% improvement over the mSDA baseline. Below, we summarize our findings when comparing the full model with the two model variants AAN-NA and AAN-NR.

\paragraph{Impact of adversarial training} We first focus on comparisons between AAN-Full and AAN-NA. The only difference between the two models is that AAN-NA has no adversarial training. On the pathology dataset, our model significantly outperforms AAN-NA, yielding a 20.2\% absolute average gain (see Table \ref{tb:pathology}). On the review dataset, our model obtains 2.5\% average improvement over AAN-NA. As shown in Table \ref{tb:review}, the gains are more significant when training on \emph{room} and \emph{checkin} aspects, reaching 6.9\% and 4.5\%, respectively. 

\paragraph{Impact of relevance scoring} As shown in Table \ref{tb:pathology}, the relevance scoring component plays a crucial role in classification on the pathology dataset. Our model achieves more than 27\% improvement over AAN-NR.
This is because in general aspects have zero correlations to each other in pathology reports. Therefore, it is essential for the model to have the capacity of distinguishing across different aspects in order to succeed in this task. 

On the review dataset, however, we observe that relevance scoring has no significant impact on performance. On average, AAN-NR actually outperforms AAN-Full by 0.9\%. This observation can be explained by the fact that different aspects in hotel reviews are highly correlated to each other. For example, the correlation between room quality and cleanliness is 0.81, much higher than aspect correlations in the pathology dataset. In other words, the sentiment is typically consistent across all sentences in a review, so that selecting aspect-specific sentences becomes unnecessary. Moreover, our supervision for the relevance scorer is weak and noisy because the aspect keywords are obtained in a semi-automatic way. Therefore, it is not surprising that AAN-NR sometimes delivers a better classification accuracy than AAN-Full.


\section{Analysis}\label{sec:analysis}

\begin{figure*}[t]
\centering
\includegraphics[width=0.95\textwidth]{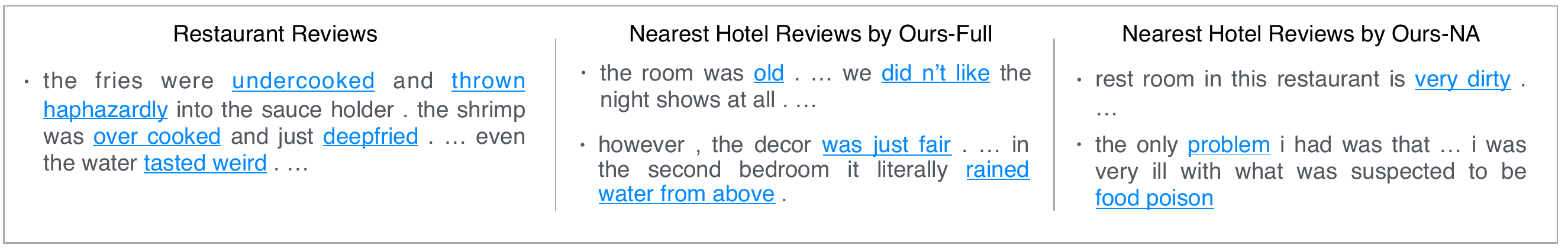}
\caption{Examples of restaurant reviews and their nearest neighboring hotel reviews induced by different models (column 2 and 3). 
We use room quality as the source aspect.
The sentiment phrases of each review are in blue, and some reviews are also shortened for space.}
\label{fig:neighbor}
\end{figure*}

\begin{table}[t]
    \centering
    \begin{tabular}{lcccc}
	\toprule
    \multirow{2}{*}{\textsc{Dataset}} & \multicolumn{2}{c}{AAN-Full} &  \multicolumn{2}{c}{AAN-NA}\\
    \cmidrule(lr){2-3} \cmidrule(l){4-5}
     & -REC. & +REC. & -REC. & +REC.\\
    \midrule
    \textsc{Pathology} & 86.2 & 91.2 & 68.6 & 72.0 \\
    \textsc{Review} & 80.8 & 86.4 & 85.0 & 83.9 \\
    \bottomrule
    \end{tabular}
    \caption{Impact of adding the reconstruction component in the model, measured by the average accuracy on each dataset. +REC. and -REC. denote the presence and absence of the reconstruction loss, respectively. }\label{tb:recon}
\end{table}

\begin{table}[t]
    \centering
    \begin{tabular}{lccc}
	\toprule
    \textsc{Dataset} & $\lambda^t=0$ & $0<\lambda^t<\infty$ & $\lambda^t=\infty$ \\
    \midrule
    \textsc{Pathology} & 77.4 & 91.2 & 81.4 \\
    \textsc{Review} & 80.9 & 86.4 & 84.3 \\
    \bottomrule
    \end{tabular}
    \caption{The effect of regularization of the transformation layer $\lambda^t$ on the performance. 
    }\label{tb:transformation}
\end{table}

\paragraph{Impact of the reconstruction loss} Table \ref{tb:recon} summarizes the impact of the reconstruction loss on the model performance. For our full model (AAN-Full), adding the reconstruction loss yields an average of 5.0\% gain on the pathology dataset and 5.6\% on the review dataset.

To analyze the reasons behind this difference, consider 
Figure \ref{fig:heat} that shows the heat maps of the learned document representations on the review dataset.
The top half of the matrices corresponds to input documents from the source domain and the bottom half corresponds to the target domain. Unlike the first matrix, the other two matrices have no significant difference between the two halves, indicating that adversarial training helps learning of domain-invariant representations. However, adversarial training also removes a lot of information from representations, as the second matrix is much more sparse than the first one. 
The third matrix shows that adding reconstruction loss effectively addresses this sparsity issue. Almost 85\% entries of the second matrix have small values ($<10^{-6}$) while the sparsity is only about 30\% for the third one. Moreover, the standard deviation of the third matrix is also ten times higher than the second one. These comparisons demonstrate that the reconstruction loss function improves both the richness and diversity of the learned representations. 
Note that in the case of no adversarial training (AAN-NA), adding the reconstruction component has no clear effect. This is expected because the main motivation of adding this component is to achieve a more robust adversarial training.


\paragraph{Regularization on the transformation layer} 
Table \ref{tb:transformation} shows the averaged accuracy with different regularization weights $\lambda^{t}$ in Equation \ref{eq:transformation}. We change  $\lambda^{t}$ to reflect different model variants. 
First, $\lambda^{t}=\infty$ corresponds to the removal of the transformation layer because the transformation is always identity in this case. Our model performs better than this variant on both datasets, yielding an average improvement of 9.8\% on the pathology dataset and 2.1\% on the review dataset. This result indicates the importance of adding the transformation layer. Second, using zero regularization ($\lambda^{t}=0$) also consistently results in inferior performance, such as 13.8\% loss on the pathology dataset. We hypothesize that zero regularization will dilute the effect from reconstruction because there is too much flexibility in transformation. As a result, the transformed representation will become sparse due to the adversarial training, leading to a performance loss.


\paragraph{Examples of neighboring reviews} Finally, we illustrate in Figure \ref{fig:neighbor} a case study on the characteristics of learned abstract representations by different models. The first column shows an example restaurant review. Sentiment phrases in this example are mostly food-specific, such as ``undercooked'' and ``tasted weird''. In the other two columns, we show example hotel reviews that are nearest neighbors to the restaurant reviews, measured by cosine similarity between their representations. In column 2, many sentiment phrases are specific for room quality, such as ``old'' and ``rained water from above''. In column 3, however, most sentiment phrases are either common sentiment expressions (e.g. dirty) or food-related (e.g. food poison), even though the focus of the reviews is based on the room quality of hotels. This observation indicates that adversarial training (AAN-Full) successfully learns to eliminate domain-specific information and to map those domain-specific words into similar domain-invariant representations. In contrast, AAN-NA only captures domain-invariant features from phrases that commonly present in both domains.

\begin{figure}[t]
\centering
\includegraphics[width=0.46\textwidth]{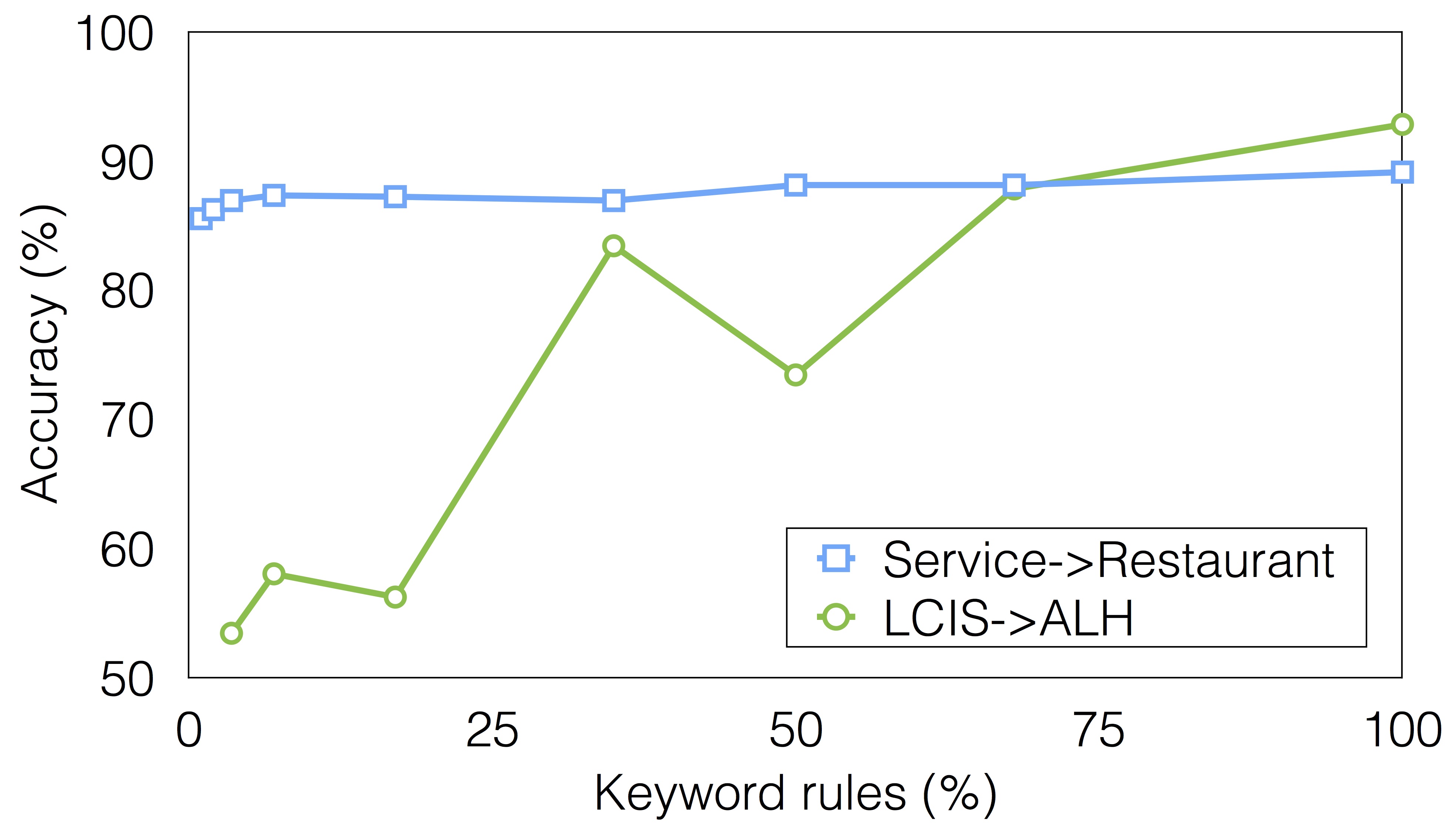}
\caption{Classification accuracy (y-axis) on two transfer scenarios (one on review and one on pathology dataset) with a varied number of  keyword rules for learning sentence relevance (x-axis).
}\label{fig:keyword}
\end{figure}

\paragraph{Impact of keyword rules} Finally, Figure \ref{fig:keyword} shows the accuracy of our full model (y-axis) when trained with various amount of keyword rules for relevance learning (x-axis). As expected, the transfer accuracy drops significantly when using fewer rules on the pathology dataset (LCIS as source and ALH as target). In contrary, the accuracy on the review dataset (hotel service as source and restaurant as target) is not sensitive to the amount of used relevance rules. This can be explained by the observation from Table \ref{tb:review} that the model without relevance scoring performs equally well as the full model due to the tight dependence in aspect labels.


%% file: conclusion.tex
\section{Conclusions}

In this paper, we propose a novel aspect-augmented adversarial network for cross-aspect and cross-domain adaptation tasks.
Experimental results demonstrate that our approach successfully learns invariant representation from aspect-relevant fragments, yielding significant improvement over the mSDA baseline and our model variants.
The effectiveness of our approach suggests the potential application of adversarial networks to a broader range of NLP tasks for improved representation learning, such as machine translation and language generation.

%% file: paper.bbl
\begin{thebibliography}{}

\bibitem[\protect\citename{Bahdanau \bgroup et al.\egroup
  }2014]{bahdanau2014neural}
Dzmitry Bahdanau, Kyunghyun Cho, and Yoshua Bengio.
\newblock 2014.
\newblock Neural machine translation by jointly learning to align and
  translate.
\newblock {\em arXiv preprint arXiv:1409.0473}.

\bibitem[\protect\citename{Blitzer \bgroup et al.\egroup
  }2006]{blitzer2006domain}
John Blitzer, Ryan McDonald, and Fernando Pereira.
\newblock 2006.
\newblock Domain adaptation with structural correspondence learning.
\newblock In {\em Proceedings of the 2006 conference on empirical methods in
  natural language processing}, pages 120--128. Association for Computational
  Linguistics.

\bibitem[\protect\citename{Blitzer \bgroup et al.\egroup
  }2007]{blitzer2007biographies}
John Blitzer, Mark Dredze, Fernando Pereira, et~al.
\newblock 2007.
\newblock Biographies, bollywood, boom-boxes and blenders: Domain adaptation
  for sentiment classification.
\newblock In {\em ACL}, volume~7, pages 440--447.

\bibitem[\protect\citename{Bordes \bgroup et al.\egroup }2012]{bordes2012joint}
Antoine Bordes, Xavier Glorot, Jason Weston, and Yoshua Bengio.
\newblock 2012.
\newblock Joint learning of words and meaning representations for open-text
  semantic parsing.
\newblock In {\em AISTATS}, volume~22, pages 127--135.

\bibitem[\protect\citename{Bousmalis \bgroup et al.\egroup
  }2016]{domain_separation_nets16}
Konstantinos Bousmalis, George Trigeorgis, Nathan Silberman, Dilip Krishnan,
  and Dumitru Erhan.
\newblock 2016.
\newblock Domain separation networks.
\newblock In {\em Neural Information Processing Systems (NIPS)}.

\bibitem[\protect\citename{Brun \bgroup et al.\egroup
  }2016]{brun-perez-roux:2016:SemEval}
Caroline Brun, Julien Perez, and Claude Roux.
\newblock 2016.
\newblock {XRCE} at {SemEval}-2016 task 5: Feedbacked ensemble modeling on
  syntactico-semantic knowledge for aspect based sentiment analysis.
\newblock In {\em Proceedings of the 10th International Workshop on Semantic
  Evaluation (SemEval-2016)}, pages 277--281.

\bibitem[\protect\citename{Buckley \bgroup et al.\egroup
  }2012]{buckley2012feasibility}
Julliette~M Buckley, Suzanne~B Coopey, John Sharko, Fernanda Polubriaginof,
  Brian Drohan, Ahmet~K Belli, Elizabeth~MH Kim, Judy~E Garber, Barbara~L
  Smith, Michele~A Gadd, et~al.
\newblock 2012.
\newblock The feasibility of using natural language processing to extract
  clinical information from breast pathology reports.
\newblock {\em Journal of pathology informatics}, 3(1):23.

\bibitem[\protect\citename{Caruana}1998]{caruana1998multitask}
Rich Caruana.
\newblock 1998.
\newblock Multitask learning.
\newblock In {\em Learning to learn}, pages 95--133. Springer.

\bibitem[\protect\citename{Chang \bgroup et al.\egroup }2007]{chang2007guiding}
Ming-Wei Chang, Lev Ratinov, and Dan Roth.
\newblock 2007.
\newblock Guiding semi-supervision with constraint-driven learning.
\newblock In {\em Annual Meeting-Association for Computational Linguistics},
  volume~45, page 280.

\bibitem[\protect\citename{Chen \bgroup et al.\egroup
  }2012]{chen2012marginalized}
Minmin Chen, Zhixiang Xu, Kilian Weinberger, and Fei Sha.
\newblock 2012.
\newblock Marginalized denoising autoencoders for domain adaptation.
\newblock {\em arXiv preprint arXiv:1206.4683}.

\bibitem[\protect\citename{Chen \bgroup et al.\egroup }2015]{chen2015abc}
Kan Chen, Jiang Wang, Liang-Chieh Chen, Haoyuan Gao, Wei Xu, and Ram Nevatia.
\newblock 2015.
\newblock Abc-cnn: An attention based convolutional neural network for visual
  question answering.
\newblock {\em arXiv preprint arXiv:1511.05960}.

\bibitem[\protect\citename{Cheng \bgroup et al.\egroup }2016]{cheng2016long}
Jianpeng Cheng, Li~Dong, and Mirella Lapata.
\newblock 2016.
\newblock Long short-term memory-networks for machine reading.
\newblock {\em arXiv preprint arXiv:1601.06733}.

\bibitem[\protect\citename{Chopra \bgroup et al.\egroup }2013]{chopra2013dlid}
Sumit Chopra, Suhrid Balakrishnan, and Raghuraman Gopalan.
\newblock 2013.
\newblock Dlid: Deep learning for domain adaptation by interpolating between
  domains.
\newblock In {\em ICML Workshop on Challenges in Representation Learning}.

\bibitem[\protect\citename{Collobert and Weston}2008]{collobert2008unified}
Ronan Collobert and Jason Weston.
\newblock 2008.
\newblock A unified architecture for natural language processing: Deep neural
  networks with multitask learning.
\newblock In {\em Proceedings of the 25th international conference on Machine
  learning}, pages 160--167. ACM.

\bibitem[\protect\citename{Eisenstein}2017]{eisenstein2017unsupervised}
Jacob Eisenstein.
\newblock 2017.
\newblock Unsupervised learning for lexicon-based classification.
\newblock In {\em {Proceedings of the National Conference on Artificial
  Intelligence (AAAI)}}.

\bibitem[\protect\citename{Ganin and Lempitsky}2014]{ganin2014unsupervised}
Yaroslav Ganin and Victor Lempitsky.
\newblock 2014.
\newblock Unsupervised domain adaptation by backpropagation.
\newblock {\em arXiv preprint arXiv:1409.7495}.

\bibitem[\protect\citename{Ganin \bgroup et al.\egroup }2015]{ganin2015domain}
Yaroslav Ganin, Evgeniya Ustinova, Hana Ajakan, Pascal Germain, Hugo
  Larochelle, Fran{\c{c}}ois Laviolette, Mario Marchand, and Victor Lempitsky.
\newblock 2015.
\newblock Domain-adversarial training of neural networks.
\newblock {\em arXiv preprint arXiv:1505.07818}.

\bibitem[\protect\citename{Glorot \bgroup et al.\egroup
  }2011]{glorot2011domain}
Xavier Glorot, Antoine Bordes, and Yoshua Bengio.
\newblock 2011.
\newblock Domain adaptation for large-scale sentiment classification: A deep
  learning approach.
\newblock In {\em Proceedings of the 28th International Conference on Machine
  Learning (ICML-11)}, pages 513--520.

\bibitem[\protect\citename{Goodfellow \bgroup et al.\egroup
  }2014]{goodfellow2014generative}
Ian Goodfellow, Jean Pouget-Abadie, Mehdi Mirza, Bing Xu, David Warde-Farley,
  Sherjil Ozair, Aaron Courville, and Yoshua Bengio.
\newblock 2014.
\newblock Generative adversarial nets.
\newblock In {\em Advances in Neural Information Processing Systems}, pages
  2672--2680.

\bibitem[\protect\citename{Grenager \bgroup et al.\egroup
  }2005]{grenager2005unsupervised}
Trond Grenager, Dan Klein, and Christopher~D Manning.
\newblock 2005.
\newblock Unsupervised learning of field segmentation models for information
  extraction.
\newblock In {\em Proceedings of the 43rd annual meeting on association for
  computational linguistics}, pages 371--378. Association for Computational
  Linguistics.

\bibitem[\protect\citename{Haghighi and Klein}2006]{haghighi2006prototype}
Aria Haghighi and Dan Klein.
\newblock 2006.
\newblock Prototype-driven learning for sequence models.
\newblock In {\em Proceedings of the main conference on Human Language
  Technology Conference of the North American Chapter of the Association of
  Computational Linguistics}, pages 320--327. Association for Computational
  Linguistics.

\bibitem[\protect\citename{Ioffe and Szegedy}2015]{ioffe2015batch}
Sergey Ioffe and Christian Szegedy.
\newblock 2015.
\newblock Batch normalization: Accelerating deep network training by reducing
  internal covariate shift.
\newblock {\em arXiv preprint arXiv:1502.03167}.

\bibitem[\protect\citename{Jagarlamudi \bgroup et al.\egroup
  }2012]{jagarlamudi2012incorporating}
Jagadeesh Jagarlamudi, Hal Daum{\'e}~III, and Raghavendra Udupa.
\newblock 2012.
\newblock Incorporating lexical priors into topic models.
\newblock In {\em Proceedings of the 13th Conference of the European Chapter of
  the Association for Computational Linguistics}, pages 204--213. Association
  for Computational Linguistics.

\bibitem[\protect\citename{Lei \bgroup et al.\egroup
  }2016]{lei2016rationalizing}
Tao Lei, Regina Barzilay, and Tommi Jaakkola.
\newblock 2016.
\newblock Rationalizing neural predictions.
\newblock {\em arXiv preprint arXiv:1606.04155}.

\bibitem[\protect\citename{Li \bgroup et al.\egroup }2012]{li2012wiki}
Shen Li, Joao~V Gra{\c{c}}a, and Ben Taskar.
\newblock 2012.
\newblock Wiki-ly supervised part-of-speech tagging.
\newblock In {\em Proceedings of the 2012 Joint Conference on Empirical Methods
  in Natural Language Processing and Computational Natural Language Learning},
  pages 1389--1398. Association for Computational Linguistics.

\bibitem[\protect\citename{Liu \bgroup et al.\egroup
  }2015]{liu2015representation}
Xiaodong Liu, Jianfeng Gao, Xiaodong He, Li~Deng, Kevin Duh, and Ye-Yi Wang.
\newblock 2015.
\newblock Representation learning using multi-task deep neural networks for
  semantic classification and information retrieval.
\newblock In {\em HLT-NAACL}, pages 912--921.

\bibitem[\protect\citename{Long \bgroup et al.\egroup
  }2016]{long2016unsupervised}
Mingsheng Long, Jianmin Wang, and Michael~I Jordan.
\newblock 2016.
\newblock Unsupervised domain adaptation with residual transfer networks.
\newblock {\em arXiv preprint arXiv:1602.04433}.

\bibitem[\protect\citename{Makhzani \bgroup et al.\egroup
  }2015]{makhzani2015adversarial}
Alireza Makhzani, Jonathon Shlens, Navdeep Jaitly, and Ian Goodfellow.
\newblock 2015.
\newblock Adversarial autoencoders.
\newblock {\em arXiv preprint arXiv:1511.05644}.

\bibitem[\protect\citename{Mann and McCallum}2008]{mann2008generalized}
Gideon~S Mann and Andrew McCallum.
\newblock 2008.
\newblock Generalized expectation criteria for semi-supervised learning of
  conditional random fields.

\bibitem[\protect\citename{Marshall \bgroup et al.\egroup
  }2015]{marshall2015robotreviewer}
Iain~J Marshall, Jo{\"e}l Kuiper, and Byron~C Wallace.
\newblock 2015.
\newblock Robotreviewer: evaluation of a system for automatically assessing
  bias in clinical trials.
\newblock {\em Journal of the American Medical Informatics Association}, page
  ocv044.

\bibitem[\protect\citename{Martins and Astudillo}2016]{martins2016softmax}
Andr{\'e}~FT Martins and Ram{\'o}n~Fernandez Astudillo.
\newblock 2016.
\newblock From softmax to sparsemax: A sparse model of attention and
  multi-label classification.
\newblock {\em arXiv preprint arXiv:1602.02068}.

\bibitem[\protect\citename{Pan and Yang}2010]{pan2010survey}
Sinno~Jialin Pan and Qiang Yang.
\newblock 2010.
\newblock A survey on transfer learning.
\newblock {\em IEEE Transactions on knowledge and data engineering},
  22(10):1345--1359.

\bibitem[\protect\citename{Pantanowitz \bgroup et al.\egroup
  }2008]{pantanowitz2008informatics}
Liron Pantanowitz, Maryanne Hornish, and Robert~A Goulart.
\newblock 2008.
\newblock Informatics applied to cytology.
\newblock {\em Cytojournal}, 5.

\bibitem[\protect\citename{Radford \bgroup et al.\egroup
  }2015]{radford2015unsupervised}
Alec Radford, Luke Metz, and Soumith Chintala.
\newblock 2015.
\newblock Unsupervised representation learning with deep convolutional
  generative adversarial networks.
\newblock {\em arXiv preprint arXiv:1511.06434}.

\bibitem[\protect\citename{Rush \bgroup et al.\egroup }2015]{rush2015neural}
Alexander~M Rush, Sumit Chopra, and Jason Weston.
\newblock 2015.
\newblock A neural attention model for abstractive sentence summarization.
\newblock {\em arXiv preprint arXiv:1509.00685}.

\bibitem[\protect\citename{Sener \bgroup et al.\egroup
  }2016]{sener2016learning}
Ozan Sener, Hyun~Oh Song, Ashutosh Saxena, and Silvio Savarese.
\newblock 2016.
\newblock Learning transferrable representations for unsupervised domain
  adaptation.
\newblock In {\em Advances In Neural Information Processing Systems}, pages
  2110--2118.

\bibitem[\protect\citename{Shinohara}2016]{shinohara2016adversarial}
Yusuke Shinohara.
\newblock 2016.
\newblock Adversarial multi-task learning of deep neural networks for robust
  speech recognition.
\newblock {\em Interspeech 2016}, pages 2369--2372.

\bibitem[\protect\citename{Springenberg}2015]{springenberg2015unsupervised}
Jost~Tobias Springenberg.
\newblock 2015.
\newblock Unsupervised and semi-supervised learning with categorical generative
  adversarial networks.
\newblock {\em arXiv preprint arXiv:1511.06390}.

\bibitem[\protect\citename{Taigman \bgroup et al.\egroup
  }2016]{taigman2016unsupervised}
Yaniv Taigman, Adam Polyak, and Lior Wolf.
\newblock 2016.
\newblock Unsupervised cross-domain image generation.
\newblock {\em arXiv preprint arXiv:1611.02200}.

\bibitem[\protect\citename{Tzeng \bgroup et al.\egroup }2014]{tzeng2014deep}
Eric Tzeng, Judy Hoffman, Ning Zhang, Kate Saenko, and Trevor Darrell.
\newblock 2014.
\newblock Deep domain confusion: Maximizing for domain invariance.
\newblock {\em arXiv preprint arXiv:1412.3474}.

\bibitem[\protect\citename{Wang \bgroup et al.\egroup }2010]{wang2010latent}
Hongning Wang, Yue Lu, and Chengxiang Zhai.
\newblock 2010.
\newblock Latent aspect rating analysis on review text data: a rating
  regression approach.
\newblock In {\em Proceedings of the 16th ACM SIGKDD international conference
  on Knowledge discovery and data mining}, pages 783--792. ACM.

\bibitem[\protect\citename{Wang \bgroup et al.\egroup }2011]{wang2011latent}
Hongning Wang, Yue Lu, and ChengXiang Zhai.
\newblock 2011.
\newblock Latent aspect rating analysis without aspect keyword supervision.
\newblock In {\em Proceedings of the 17th ACM SIGKDD international conference
  on Knowledge discovery and data mining}, pages 618--626. ACM.

\bibitem[\protect\citename{Xu and Saenko}2015]{xu2015ask}
Huijuan Xu and Kate Saenko.
\newblock 2015.
\newblock Ask, attend and answer: Exploring question-guided spatial attention
  for visual question answering.
\newblock {\em arXiv preprint arXiv:1511.05234}.

\bibitem[\protect\citename{Xu \bgroup et al.\egroup }2015]{xu2015show}
Kelvin Xu, Jimmy Ba, Ryan Kiros, Kyunghyun Cho, Aaron Courville, Ruslan
  Salakhutdinov, Richard~S Zemel, and Yoshua Bengio.
\newblock 2015.
\newblock Show, attend and tell: Neural image caption generation with visual
  attention.
\newblock {\em arXiv preprint arXiv:1502.03044}, page~5.

\bibitem[\protect\citename{Yala \bgroup et al.\egroup }2016]{Yala:2016a}
Adam Yala, Regina Barzilay, Laura Salama, Molly Griffin, Grace Sollender,
  Aditya Bardia, Constance Lehman, Julliette~M Buckley, Suzanne~B Coopey,
  Fernanda Polubriaginof, J~Garber, BL~Smith, MA~Gadd, MC~Specht, and
  TM~Gudewicz.
\newblock 2016.
\newblock Using machine learning to parse breast pathology reports.
\newblock {\em Breast Cancer Research and Treatment}.

\bibitem[\protect\citename{Yang \bgroup et al.\egroup }2015]{yang2015stacked}
Zichao Yang, Xiaodong He, Jianfeng Gao, Li~Deng, and Alex Smola.
\newblock 2015.
\newblock Stacked attention networks for image question answering.
\newblock {\em arXiv preprint arXiv:1511.02274}.

\bibitem[\protect\citename{Zaidan \bgroup et al.\egroup }2007]{zaidan2007using}
Omar Zaidan, Jason Eisner, and Christine~D Piatko.
\newblock 2007.
\newblock Using ``annotator rationales'' to improve machine learning for text
  categorization.
\newblock In {\em HLT-NAACL}, pages 260--267. Citeseer.

\bibitem[\protect\citename{Zhang \bgroup et al.\egroup
  }2016]{zhang2016rationale}
Ye~Zhang, Iain Marshall, and Byron~C Wallace.
\newblock 2016.
\newblock Rationale-augmented convolutional neural networks for text
  classification.
\newblock {\em arXiv preprint arXiv:1605.04469}.

\bibitem[\protect\citename{Zhou \bgroup et al.\egroup }2016]{zhou2016bi}
Guangyou Zhou, Zhiwen Xie, Jimmy~Xiangji Huang, and Tingting He.
\newblock 2016.
\newblock Bi-transferring deep neural networks for domain adaptation.
\newblock ACL.

\end{thebibliography}
